\documentclass{article}
\usepackage{spconf,amsmath,graphicx}
\usepackage{multirow}
\usepackage{array}
\usepackage{float}
\usepackage{placeins}
\usepackage{varioref}
\usepackage{setspace}
\usepackage{color, colortbl}
\usepackage{enumitem}
\usepackage{kotex}
\usepackage{tikz}
\usepackage{caption}
\def\checkmark{\tikz\fill[scale=0.4](0,.35) -- (.25,0) -- (1,.7) -- (.25,.15) -- cycle;} 

\setlist[itemize]{align=parleft,left=0pt..1.2em}
\setlist[enumerate]{align=parleft,left=0pt..1.2em}

\title{Improving Robustness of Neural Inverse Text Normalization via Data-Augmentation, Semi-Supervised Learning, and Post-Aligning Method}
\name{Juntae Kim, Minkyu Lim, and Seokjin Hong}
\address{SK Telecom, Seoul, Republic of Korea}
\begin{document}
\ninept
\maketitle
\begin{abstract}
Inverse text normalization (ITN) is crucial for converting spoken-form into written-form, especially in the context of automatic speech recognition (ASR). While most downstream tasks of ASR rely on written-form, ASR systems often output spoken-form, highlighting the necessity for robust ITN in product-level ASR-based applications. Although neural ITN methods have shown promise, they still encounter performance challenges, particularly when dealing with ASR-generated spoken text. These challenges arise from the out-of-domain problem between training data and ASR-generated text. To address this, we propose a direct training approach that utilizes ASR-generated written or spoken text, with pairs augmented through ASR linguistic context emulation and a semi-supervised learning method enhanced by a large language model, respectively. Additionally, we introduce a post-aligning method to manage unpredictable errors, thereby enhancing the reliability of ITN. Our experiments show that our proposed methods remarkably improved ITN performance in various ASR scenarios.

\end{abstract}
\begin{keywords}
ASR, inverse text normalization
\end{keywords}
\section{Introduction}
\label{sec:intro}


Inverse text normalization (ITN) converts spoken-form text to its corresponding written-form, which is used as post-processing for an automatic speech recognition (ASR) system's output since it typically provides spoken-form text. While some ASR systems \cite{radford2023robust} can generate written-form text through training with written-form text data, they may still occasionally produce spoken-form text, resulting in unstable conversion performance compared to applying ITN. Written-form text is preferred due to its user-friendliness, and many downstream tasks of ASR rely on written-form text, such as date-based scheduling and address-based search. Thus, ensuring robust ITN performance is essential for a product-level ASR system \cite{9414912}.

Recently, neural ITN methods \cite{9414912, 10094599, antonova22_interspeech, 10023257, paul2022improving} have outperformed the previous finite-state transducer (FST)-based ITN approaches \cite{shugrina2010formatting, zhang21ja_interspeech, ebden2015kestrel}. This is because leveraging the data-driven modeling capabilities of neural networks is more advantageous for capturing numerous spoken-written form relationships than FST, which incurs expensive scaling costs for transformation rules within the FST graph.



Despite promising results in neural ITN, significant performance degradation occurs when applying it to ASR-generated spoken text \cite{9414912, 10094599}. Neural ITN is trained on the general spoken text and its corresponding written text, which may lead to differences in linguistic context compared to ASR-generated spoken text due to irregular interjections and ASR errors, i.e., the performance degradation stems from the out-of-domain problem, indicating that addressing this issue solely through an ordinary text corpus has limited effectiveness. To resolve the out-of-domain problem, the most direct approach is to utilize ASR-generated spoken or written text and their corresponding pairs for training neural ITN. However, constructing such a corpus with human resources is costly, making it nearly impractical to scale up the amount of ASR spoken-written text in this manner.


Furthermore, neural ITN methods based on sequence-to-sequence (seq2seq) models \cite{9414912, ihori2020large} can introduce unpredictable errors, which pose a critical challenge for product-level solutions as these errors may distort the original meaning of ASR-generated spoken text. To address this issue, \cite{10094599, antonova22_interspeech} proposed an additional tagger network to specify the normalization region; however, since the tagger network still relies on a neural network-based approach, there is no guarantee to control unpredictable errors associated with neural networks.


In light of the aforementioned problems, this paper contributes the following:
\begin{itemize}


\item To address the out-of-domain problem, we directly train neural ITN using ASR-generated text data. We create spoken-written text pairs from ASR-generated spoken or written text using a data augmentation (DA) and semi-supervised learning (SSL) pipeline. DA emulates the linguistic context of ASR-generated spoken text to enhance the robustness of neural ITN, while SSL employs a large language model (LLM) for confidence scoring to prevent neural ITN from learning inaccuracies from pseudo-written text.



\item To address neural ITN's unpredictable errors, we introduce a post-aligning (PA) algorithm that provides better control than neural network-based models alone. PA detects discrepancies between spoken and neural ITN hypotheses, making targeted replacements. It also utilizes all neural ITN hypotheses, not just the top-ranked one, for enhanced performance.

\end{itemize}
Experiments exhibit that applying our proposed methods showed remarkable robustness in various ASR scenarios. Although our experiments were done in Korean language, most proposed methods can be extended to other languages.

\section{Data Generation Pipeline}
\label{sec:format}
In Fig. 1, we illustrate the proposed training data generation pipeline. To train the neural ITN, spoken-written form text pairs are essential. However, we can obtain either spoken or written text from the ASR system. The objective of our pipeline is to create pairs solely based on spoken or written text. To accomplish this, in a high-level description, DA generates spoken text from written text, and vice versa is achieved by SSL; thus, we achieve the training of the neural ITN with ASR-generated text. Consequently, the pipeline is applied to the three types of datasets: spoken-written text corpus (type I), written text corpus (type II), and spoken text corpus (type III), where type I is from text corpus for Korean text normalization publicly released by AI Hub\footnote{https://www.aihub.or.kr}; type II and type III are from our in-house ASR system trained using both spoken and written ASR labels.

\subsection{Data augmentation}
\label{ssec:subhead}

\begin{figure}[t]
  \centering
  \includegraphics[width=\linewidth]{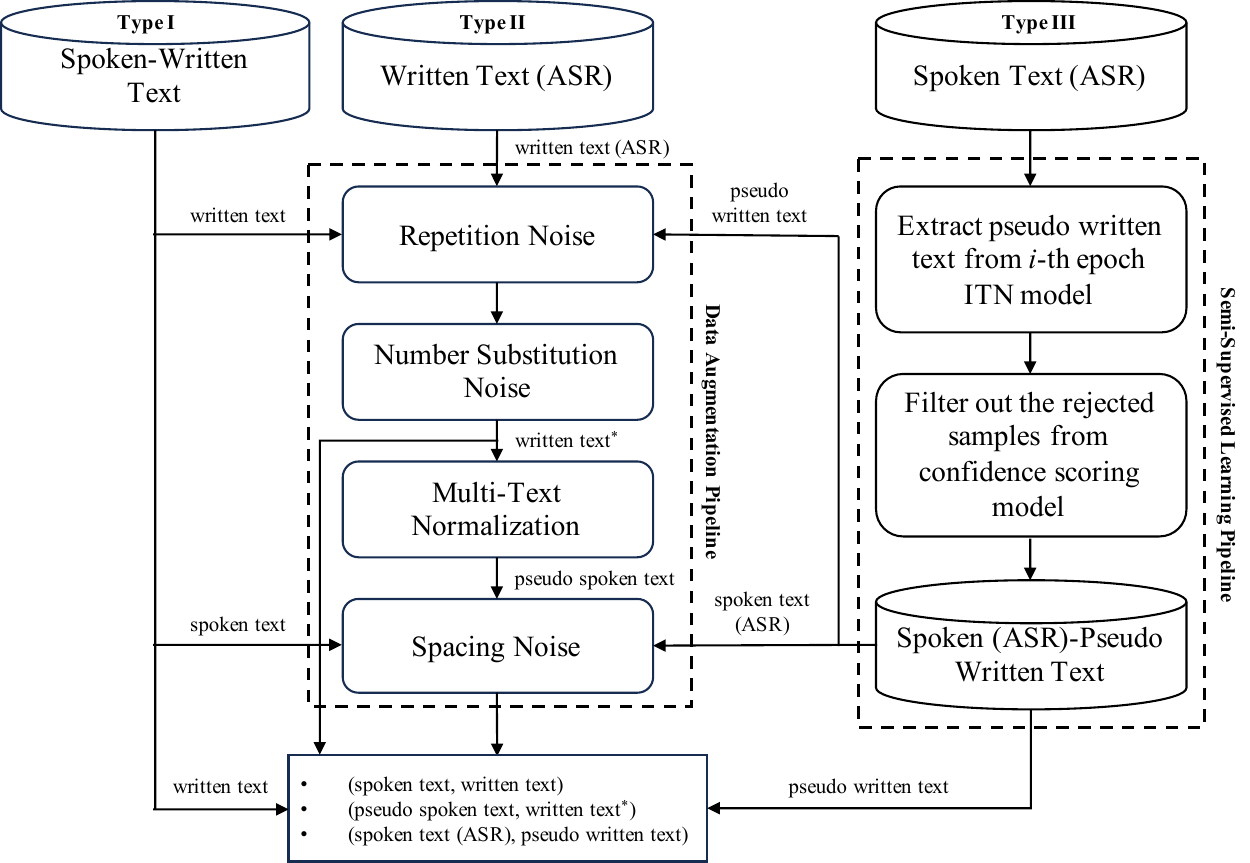}
  \caption{Training data generation pipeline with DA and SSL based on dataset types.}
  \vspace{-0.2cm}

\end{figure}

As shown in Fig. 1, DA pipeline includes noise augmentation and multi-text normalization (MTN). Given written text, DA generates the noise-augmented written text ($\text{written text}^{*}$) and its corresponding spoken text (pseudo-spoken text). The used noise types in Fig. 1 are chosen to compensate for the weak points of the baseline neural ITN, when it is exposed to ASR spoken text. The baseline neural ITN was trained using type I corpus alone without any ASR-related datasets. Note that widely used noise types for DA in the general natural language processing field \cite{wei2019eda} was not effective according to our experiments.

To generate $\text{written text}^{*}$, repetition noise is added to written text at first. The repetition is common in ASR spoken text because a speaker can actually repeat the words or ASR system can cause the repetition error. Regardless of where this error originated from, the baseline neural ITN showed the vulnerability when it was exposed by the repetition error. For example, baseline neural ITN repeats the words more than the spoken text actually did. To mitigate this case by training, we randomly repeat the words in the given written text. After that, the numbers in the written text are substituted into other random numbers to expand the coverage of numbers.

To generate spoken text given $\text{written text}^{*}$, MTN is adopted. The spoken-written form has a many-to-one relationship rather than one-to-one. For example, if the written form is ``2024", people can say ``two oh two four", ``twenty twenty-four", or ``two thousand twenty-four". However, the common text normalization techniques only support the one-to-one relationship. To address this problem, we expand the text normalization to MTN to support a many-to-one relationship. MTN randomly outputs the spoken form among the candidates given written form and we refer the output of MTN to pseudo-spoken text.

In Korean, spacing rules in writing are considered unstable, resulting in inconsistent adherence by native speakers \cite{kim2013keeping}. Consequently, the Korean ASR corpus contains numerous spacing errors, indicating the frequent occurrence of spacing errors caused by the Korean ASR system. For example, despite one of the spoken forms of ``2013" being ``이천십삼", the ASR system can produce variations such as ``이천 십삼", ``이 천십삼", or ``이천십 삼". To replicate this effect, the DA pipeline includes spacing noise augmentation applied to spoken-form text. Spacing errors are introduced exclusively in regions that deviate from the written form to prevent over-regularization.

To sum up, DA generates a pair of pseudo-spoken text and $\text{written text}^{*}$ from written text. Note that the written text can come from type I and III as well as type II as illustrated in Fig. 1, in order to effectively expand the amount of training data. Also, spacing noise augmentation is applied to all spoken text regardless of the type of the dataset.

\subsection{Semi-Supervised learning}
\label{ssec:subhead}

Our SSL pipeline in Fig. 1 comprises four primary steps \cite{xie2020self}:
\begin{enumerate}
\item Training a teacher ITN model on type I and II datasets with DA.
\item Generating pseudo-written text using the teacher model on the type III dataset.
\item Filtering out some of the pseudo-written text based on confidence scoring.
\item Training a student model on all spoken-written text pairs generated from the data generation pipeline: (spoken text, written text), (pseudo-spoken text, $\text{written text}^{*}$), and (spoken text (ASR), pseudo-written text).
\end{enumerate}

\begin{figure*}[t]
  \centering
  \includegraphics[width=\linewidth]{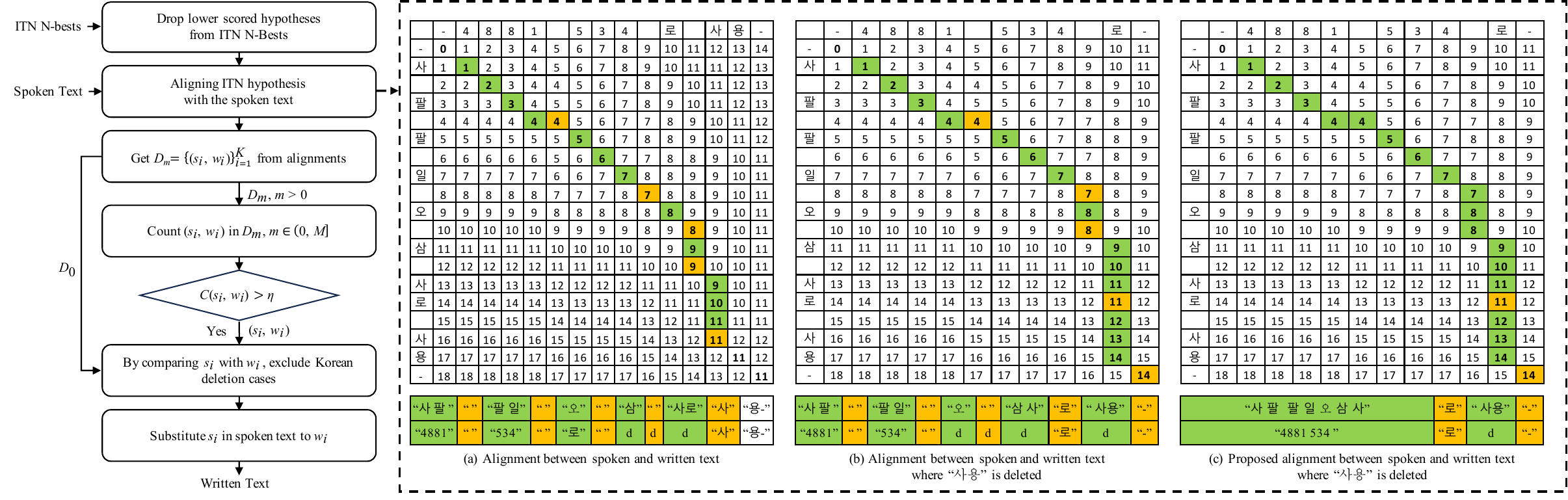}
  \caption{PA block diagram and examples of alignments between spoken-text (column) and written-text (row). The bold blue numbers indicate the alignment path, orange boxes represent boundary syllables, and green boxes highlight normalization regions.}
\vspace{-0.2cm}

\end{figure*}

We iterate through this procedure over multiple epochs, wherein the trained student at each epoch serves as a teacher to generate new pseudo-written text from the spoken text and train a new student. Note that DA is also applied to the pseudo-written text generated by the SSL pipeline. This process encourages the student model to learn a more fluent context of ASR text than the teacher model initially had from the type III dataset.

Confidence scoring plays a crucial role in SSL \cite{yao2022enhancing}. The traditional approach to confidence scoring relies on the likelihood of the teacher model for unlabeled data \cite{lee2013pseudo}. However, SSL using this approach did not demonstrate improvement in our work because the likelihood measure fails to discern unnaturally converted pseudo-written text. For instance, consider the sentence: ``응 응 오늘 아 할 일 많아" (``Yes, yes, today, there is a lot of work to do"); neural ITN may erroneously convert it to ``응 응 오늘 아 할 1 많아", as the Korean word ``일" can mean both ``work" and ``one". However, ``응 응 오늘 아 할 1 많아" is linguistically unnatural because ``일" is intended to mean ``work" in this context. Unfortunately, likelihood-based scoring methods occasionally assign a high score to such unnatural sentences.

The ideal confidence scoring method for ITN should assess the naturalness of written text converted by neural ITN to obtain reliable pseudo-written text. However, developing a model to evaluate naturalness for confidence scoring is not a trivial task. Instead, we assume that recently proposed approaches based on LLMs \cite{ouyang2022training, min2021recent, brown2020language} have the potential to assess naturalness effectively. Considering that LLMs are trained on a vast amount of natural language data, and naturalness can be defined by expressions commonly used by people, LLMs have the potential to discriminate linguistically unnatural sentences.

To verify this assumption, we adopt Chat-GPT\footnote{https://chat.openai.com}, one of the representative LLMs, with the following prompts:

\begin{itemize}
    \item Which of the two sentences, \{spoken text\} or \{pseudo-written text\} when the numeric part of \{spoken text\} is converted to numbers, is more commonly used by people? Do you think people would find it easier to recognize if the altered sentence is quoted in a news article? Please respond with ``yes" or ``no" only.
    \item Tell me if \{pseudo-written text\} is the correct conversion for the numeric part pronounced in Korean in \{spoken text\}. Please respond with only ``yes" or ``no".
\end{itemize}




At the first prompt, using the second sentence enhances Chat-GPT's ability to make decisions about whether the pseudo-written text closely resembles standard written form, as news articles typically quote sentences in this form. The second prompt is similar to the first one. Since Chat-GPT's responses can vary across trials, asking Chat-GPT a similar prompt again can improve the reliability of the answer, especially if we consider the pseudo-written text as valid only when Chat-GPT responds positively to both prompts.

For the rejected samples that received a negative response from Chat-GPT to one of the prompts, we utilized them by directly copying the spoken-form text to use it as pseudo-written text instead of discarding them. While this method may not reduce quantitative errors, it can be beneficial from a qualitative perspective, as incorrect conversions can distort the original meaning of the spoken text.

\section{Post-Aligning}
\label{sec:format}

As depicted in Fig. 2, certain hypotheses are firstly dropped from $N$-bests when the difference in scores between them and the 1-best hypothesis exceeds the threshold $\alpha$ because lower-scored hypotheses usually have erroneous conversions. We assume that $M$ hypotheses are remained from this process. After that, the alignment between spoken text and each ITN hypothesis is conducted based on edit-distance. From the alignment, the set of $K$ normalization pairs, $D_m = \left\{\left(s_i, w_i\right)\right\}_{i=1}^K$ is obtained where $m$ is the hypothesis number, $s_i$ and $w_i$ are the normalization region and its target in spoken and written text, respectively. For intuitive understanding, we provide an example in Fig. 2-(a) in which the normalization regions correspond to green boxes of which the start point is found when the current score in the path is higher than the previous score and the end-point is found when the current score is the same with the previous score.

Although the written text can be recovered by replacing the $s_i$ to $w_i$ in spoken text, the baseline alignment has a limitation when a deletion error occurs in written text. In Fig. 2-(b), if we allow the deletion normalizations such as (``오", ``"), the erroneously deleted ``사용" by neural ITN cannot be recovered. The deletion normalization is mainly due to the existence of numerous spacing in Korean spoken text because the spacing characters in spoken text are matched with certain numbers in written text instead of Korean syllables, leading to deletion normalization. To cope with this problem, we proposed a blacklist passing scheme which does not allow the characters as an end-point if the characters are in the blacklist. As described in Fig. 3-(c), the proposed alignment detects actual erroneous deletion (``사용", ``"); thus, we can restore ``사용", by not allowing the deletion normalization.

Given $D_m$, we identify the candidates of normalization pairs involving the replacement. All pairs in $D_0$ are included in the candidates. The repeatedly appeared pairs in $D_1$ to $D_M$ are counted and if the counting number of a certain pair is greater than $\eta$, it is also included in the candidates. Note that the spacing in $s_i$ and $w_i$ is not considered for counting, e.g., (``사십", ``40") and (``사 십", ``40") is considered to be the same. By doing so, we expected the boosting effect of neural ITN because hypotheses more than 1-best are exploited via this process for written text prediction.

From the candidates, we ignore the Korean deletion normalization to restore the deletion errors caused by neural ITN. Although we only deal with deletion errors, other error types also can be handled by providing some rules at this stage. Finally, by substituting $s_i$ in the spoken text to $w_i$ in candidates, we can obtain the written text.     

\definecolor{brightgray}{gray}{0.9}

\begin{table*}[t]
\centering
\caption{ITN performance comparison on AI-Hub, ASR-TEL, and ASR-BC considering the application of methods (DA, SSL, and PA) and dataset utilization for training (type I, type II, and type III). The numbers in bold indicate the best result.}
\label{table}
\small
\setlength{\tabcolsep}{1pt}
\renewcommand{\arraystretch}{1.1}

\begin{tabular}{>{\centering}m{20pt}>{\centering}m{20pt}>{\centering}m{20pt}|>{\centering}m{35pt}>{\centering}m{35pt}>{\centering}m{35pt}|>{\centering}m{34pt}>{\centering}m{34pt}>{\centering}m{34pt}|>{\centering}m{34pt}>{\centering}m{34pt}>{\centering}m{34pt}|>{\centering}m{34pt}>{\centering}m{34pt}>{\centering}m{34pt}}

\noalign{\hrule height 0.8pt}

\multicolumn{3}{c|}{\textbf{Method}} & \multicolumn{3}{c|}{\textbf{Training Dataset}} & \multicolumn{ 9}{c}{\textbf{Test Dataset}} \tabularnewline
\hline
\multirow{ 2}{*}{\textbf{DA}} & \multirow{ 2}{*}{\textbf{SSL}} & \multirow{ 2}{*}{\textbf{PA}} & \multirow{ 2}{*}{\textbf{Type I}} & \multirow{ 2}{*}{\textbf{Type II}} & \multirow{ 2}{*}{\textbf{Type III}} & \multicolumn{ 3}{c|}{\textbf{AI-Hub}} & \multicolumn{ 3}{c|}{\textbf{ASR-TEL (ITN label)}} & \multicolumn{ 3}{c}{\textbf{ASR-BC (ASR label)}}\tabularnewline

\cline{7-15}

& & & & & & \textbf{I-CER} & \textbf{NI-CER} & \textbf{CER} & \textbf{I-CER} & \textbf{NI-CER} & \textbf{CER} & \textbf{I-CER} & \textbf{NI-CER} & \textbf{CER} \tabularnewline
\hline

& & & \checkmark & & & \textbf{0.32} & \textbf{0.34} & \textbf{0.34} & 49.58 & 6.00 & 6.45 & 41.54 & 21.18 & 21.40 \tabularnewline
\checkmark & & & \checkmark & & & 0.77 & 0.64 & 0.66 & 56.09 & 1.39 & 1.96 & 27.31 & 15.96 & 16.09 \tabularnewline
\checkmark & & & \checkmark & \checkmark & & 0.66 & 0.65 & 0.65 & 15.90 & 0.24 & 0.40 & 17.85 & 15.01 & 15.04\tabularnewline
\checkmark & \checkmark &  & \checkmark & \checkmark & \checkmark & 0.74 & 0.68 & 0.69 & 12.49 & 0.21 & 0.34 & 16.43 & 14.90 & 14.92 \tabularnewline
\checkmark & & \checkmark & \checkmark & & & 0.76 & 0.61 & 0.63 & 54.58 & 0.55 & 1.12 & 26.74& 14.69 & 14.82 \tabularnewline
\checkmark & & \checkmark & \checkmark & \checkmark & & 0.65 & 0.66 & 0.66 & 15.14 & \textbf{0.17} & 0.33 & 17.35 & \textbf{14.64} & 14.67 \tabularnewline

\checkmark & \checkmark & \checkmark & \checkmark & \checkmark & \checkmark & 0.73 & 0.69 & 0.70 & \textbf{11.96} & 0.18 & \textbf{0.31} & \textbf{16.36} & \textbf{14.64} & \textbf{14.66}\tabularnewline

\noalign{\hrule height 0.8pt}
\end{tabular}
\label{tab3}
\vspace{-0.2cm}

\end{table*}

\section{Experiments and Results}
\subsection{Experimental setup}

\textbf{Model:} For neural ITN, we employed a conventional transformer-based seq2seq model. The encoder and decoder networks consist of 12 and 1 layer(s), respectively, with all layers having 1024 hidden units. For beam search, we used a beam size of 5. For PA, $\alpha$ and $\eta$ were set to 5 and 1, respectively, based on our validation set. For SSL, we initially trained our neural ITN on type I and II datasets with DA until the validation loss no longer decreased. Subsequently, we conducted additional training for 10 epochs, including the SSL. As the baseline, we employed the model trained exclusively on the type I dataset without the application of DA, SSL, or PA.

In the case of ASR, we adopted the conformer-transducer model \cite{guo2021recent, kim2020accelerating}, which was trained on our in-house dataset comprising 13k hours of Korean utterances. Note that our ASR labels encompass both spoken and written forms, enabling our ASR system to generate both written and spoken outputs.

\textbf{Datasets:} For training, we utilize AI-Hub and ASR-generated datasets for written and spoken text, corresponding to type I, type II, and type III datasets, respectively. AI-Hub contains 300k spoken-written text pairs and covers a wide range of entities, including floats, times, currencies, units, addresses, and account numbers. We divided AI-Hub into subsets of 240k, 10k, and 50k sentences for training, validation, and testing, respectively. To obtain diverse ASR error distributions, we applied our ASR system to 4M utterances from various domains, including smart speakers, broadcasts, and conversational domains. Of these, 3M utterances were used to train the ASR system, while the remaining 1M were not. Among the 4M utterances, we obtained 1M in written form (type II), and we randomly extracted 1M in spoken form from the remaining 3M utterances (type III). Note that we distinguished the written form by assessing whether the ASR-generated text includes non-Korean characters.

For testing, we have adopted three test sets: the AI-Hub test set (50k sentences), ASR-generated datasets from an additional 5k utterances recorded in a phone-call scenario (ASR-TEL), and 5k utterances in a broadcasting scenario (ASR-BCN). Since using ASR labels as the ground truth text cannot measure the pure and potential performance of the neural ITN, we manually labeled the ASR-TEL for ITN evaluation while we continued to use existing ASR labels for ASR-BCN. The character error rates (CER) of our ASR system for ASR-TEL and ASR-BCN were 7.0\% and 15.89\%, respectively.

\textbf{Metrics:} The ITN performance is measured by ITN CER (I-CER), and Non-ITN CER (NI-CER), focusing on the quality of text normalization and measuring undesirable perturbations to the original source text, respectively \cite{9414912}. CER is also utilized.

\subsection{Experimental results and discussion}

Table I summarizes the experimental results, demonstrating the significant effectiveness of DA, SSL, and PA in ASR-TEL (CER: 0.31\%) and ASR-BCN (CER: 14.66\%), except for AI-Hub, where the baseline achieved the best results (CER: 0.34\%). The degradation in AI-Hub is attributed to differences in written form styles between AI-Hub and ASR-generated text datasets. For example, the spoken form ``십만원" (one hundred thousand won) can be represented in both written forms: ``10만원" or ``100,000원". Additionally, the noise augmentation methods in DA primarily aim to simulate ASR-generated text with distinct linguistic contexts compared to ordinary text. This implies that DA may have led to an over-regularization effect on the AI-Hub test set. However, it's worth noting that the I-CER, NI-CER, and CER for all proposed methods in AI-Hub remain below 1\%, which is reasonable considering the performance improvements in ASR-TEL and ASR-BCN.

Applying DA to type I resulted in a remarkable reduction in NI-CER in ASR-TEL (6.00\% $\to$ 1.39\%) and ASR-BCN (21.18\% $\to$ 15.96\%) compared to the baseline, while a reduction in I-CER was only evident in ASR-BCN (41.54\% $\to$ 27.31\%). This suggests that noise augmentation in DA effectively mitigates the distortion of the original text by the neural ITN. Furthermore, given that ASR-BCN comprises broadcasts, including news content, the semantic context for ITN in ASR-BCN can be closer to that of AI-Hub than ASR-TEL, which has a more irregular semantic context due to its origin in a fully conversational domain. This leads to different results regarding I-CER. This is supported by the results when type II, including conversational domains, is involved for training, which showed a dramatic reduction in I-CER: 15.90\% and 17.85\% in ASR-TEL and ASR-BCN, while showing different results in NI-CER: 0.24\% and 15.01\% in ASR-TEL and ASR-BCN.

When applying SSL, we observed an improvement in CER in both ASR-TEL and ASR-BCN, with marginal degradation in NI-CER. We found that this degradation depends on the reliability of the confidence-scoring model, which restricts erroneous pseudo-written text for training the student model. This is why we used two similar prompts to improve reliability.

PA consistently improved performance in both I-CER and NI-CER in ASR-TEL and ASR-BCN. PA selectively carries out ITN while conserving the original text not involved in ITN, leading to NI-CER reduction. Also, PA utilizes $M$-best hypotheses when building the ITN candidates, leading to I-CER reduction because additional ITN conversions are achieved if some hypotheses have correct predictions when the 1-best has failed.

Note that the relative performance improvements with proposed methods in CER in ASR-TEL with ITN labels compared to the baseline was 95.18\%, implying that our proposed methods have outstanding potential power for ITN if the ASR system provides correct outputs to the neural ITN.

\definecolor{brightgray}{gray}{0.9}

\section{Conclusion}
In this work, we proposed novel methods to address the performance degradation of neural ITN when it encounters ASR-generated text due to the out-of-domain problem. To enable the direct utilization of ASR-generated text for neural ITN training, DA and SSL effectively generated spoken-written text pairs from just one of them, leading to a significant improvement in performance. PA provided further enhancements through its selective and boosted ITN conversion approach. Although our datasets also include foreign words, this work primarily focuses on the conversion of numeric words. Our future work will expand to effectively handle foreign words as well.


\begin{spacing}{0.0}

\bibliographystyle{IEEEtran}
\bibliography{main}
\end{spacing}

\end{document}